\def\year{2021}\relax
\documentclass[letterpaper]{article} % DO NOT CHANGE THIS
\usepackage{aaai21}  % DO NOT CHANGE THIS
\usepackage{times}  % DO NOT CHANGE THIS
\usepackage{helvet} % DO NOT CHANGE THIS
\usepackage{courier}  % DO NOT CHANGE THIS
\usepackage[hyphens]{url}  % DO NOT CHANGE THIS
\usepackage{graphicx} % DO NOT CHANGE THIS
\usepackage{natbib}  % DO NOT CHANGE THIS AND DO NOT ADD ANY OPTIONS TO IT
\usepackage{caption} % DO NOT CHANGE THIS AND DO NOT ADD ANY OPTIONS TO IT
\usepackage[cmex10]{amsmath} 
\usepackage{amsthm}
\usepackage{amssymb}
\usepackage[ruled,algo2e]{algorithm2e}
\usepackage{algorithm}
\usepackage[mathcal]{euscript} 

\DeclareMathOperator{\cov}{cov}
\DeclareMathOperator{\tr}{tr}
\DeclareMathOperator*{\argmax}{arg\,max}

\title{A Maximal Correlation Approach to Imposing Fairness in Machine Learning }
\author{
    Joshua Lee\textsuperscript{\rm 1}, Yuheng Bu\textsuperscript{\rm 1}, Prasanna Sattigeri\textsuperscript{\rm 2}, Rameswar Panda\textsuperscript{\rm 2}, Gregory Wornell\textsuperscript{\rm 1}, Leonid Karlinsky\textsuperscript{\rm 2}, Rogerio Feris\textsuperscript{\rm 2}
    \\
}
\affiliations{
    \textsuperscript{\rm 1}Dept. EECS, MIT
    \\
    \textsuperscript{\rm 2}MIT-IBM Watson AI Lab, IBM Research
}
\begin{document}

\newcommand{\prasanna}[1]{{\textcolor{blue}{Prasanna: #1}}}
\newcommand{\joshua}[1]{{\textcolor{magenta}{Joshua: #1}}}
\newcommand{\yuheng}[1]{{\textcolor{red}{Yuheng: #1}}}

\newcommand{\HGR}{\mathrm{HGR}}
\newcommand{\soft}{\mathrm{soft}}
\newcommand{\kNN}{$k$NN }

\newcommand{\T}{\mathrm{T}}
\newcommand{\cT}{\mathcal{T}}
\newcommand{\Ph}{\hat{P}}
\newcommand{\cL}{\mathcal{L}}
\newcommand{\cX}{\mathcal{X}}
\newcommand{\cY}{\mathcal{Y}}
\newcommand{\cD}{\mathcal{D}}
\newcommand{\yh}{\hat{y}}
\newcommand{\bff}{\mathbf{f}}
\newcommand{\bg}{\mathbf{g}}
\newcommand{\bh}{\mathbf{h}}
\newcommand{\bB}{\mathbf{B}}
\newcommand{\reals}{\mathbb{R}}
\newcommand{\bI}{\mathbf{I}}
\newcommand{\bzero}{\mathbf{0}}
\newcommand{\defeq}{\triangleq}

\newcommand{\ev}{{\mathbb{E}}}
\newcommand{\pr}{\mathbb{P}}
\newcommand{\prob}[1]{\pr\left\{#1\right\}}
\newcommand{\E}[1]{\ev\left[#1\right]}
\newcommand{\bE}[1]{\ev\bigl[#1\bigr]}
\newcommand{\Eh}[2]{\hat{\ev}{}_{#1}\left[#2\right]}
\newcommand{\bEh}[2]{\hat{\ev}{}_{#1}\bigl[#2\bigr]}
\newcommand{\varh}[2]{\widehat{\var}{}_{#1}\left[#2\right]}
\newcommand{\bvarh}[2]{\widehat{\var}{}_{#1}\bigl[#2\bigr]}
\newcommand{\Ed}[2]{\ev_{#1}\left[#2\right]}
\newcommand{\bEd}[2]{\ev_{#1}\bigl[#2\bigr]}

\newtheorem{theorem}{Theorem}
\newtheorem{definition}{Definition}
\newtheorem{lemma}[theorem]{Lemma}

\maketitle

\begin{abstract}
As machine learning algorithms grow in popularity and diversify to many industries, ethical and legal concerns regarding their fairness have become increasingly relevant. We explore the problem of algorithmic fairness, taking an information-theoretic view. The maximal correlation framework is introduced for expressing fairness constraints and shown to be capable of being used to derive regularizers that enforce independence and separation-based fairness criteria, which admit optimization algorithms for both discrete and continuous variables which are more computationally efficient than existing algorithms. We show that these algorithms provide smooth performance-fairness tradeoff curves and perform competitively with state-of-the-art methods on both discrete datasets (COMPAS, Adult) and continuous datasets (Communities and Crimes).
\end{abstract}

\section{Introduction}

The use of machine learning in many industries has raised many ethical and legal concerns, especially that of fairness and bias in predictions \cite{selbst2019fairness,bellamy2018ai}. As systems are trusted to aid or make decisions regarding loan applications, criminal sentencing, and even health care, it is vital that unfair biases do not influence them.

However, mitigating these biases is complicated by ever-changing perspectives on fairness, and a good system for enforcing fairness must be adaptable to new settings. In particular, there are often competing notions on fairness. Two of these popular notions are independence and separation (a third condition, sufficiency, is beyond the scope of this paper) \cite{barocas-hardt-narayanan}. Independence ensures that predictions are independent from membership in a protected class, so that one achieves equal favourable outcome rates across all groups, and arises in applications such as affirmative action \cite{equal1978department}. Separation is designed to achieve equal type I/II error rates across all groups by enforcing independence between predictions and membership in a protected class conditional on the class label. This criterion is used to measure fairness in recidivism predictions and bank loan applications. A significant body of work \cite{locatello2019fairness, golz2019paradoxes,corbett2018measure,barocas-hardt-narayanan} has gone into explaining that independence and separation are inherently incompatible for non-trivial cases and their applicability needs to be determined by the application and the stakeholders. This motivates us to construct a framework that is flexible enough to handle different fairness criteria, and to do it with different modalities of data (discrete vs. continuous data, for example).

This bias mitigation must also be balanced out with the system's usefulness, and often one must tune the tradeoff between the fairness (as measured in the particular context) and performance according to a current situation, which can be a difficult process if the tradeoff curve is not smooth. Generating the frontier of possible values can be computationally infeasible or impossible if the algorithm does not have a regularization parameter to adjust, \cite{calmon2017optimized, mary2019fairness}, thus making it difficult to achieve this balance, which makes fast generation of fair classifiers even more important.

%have numerous advantages and disadvantages. 

%\prasanna{Should this paragraph come in later sections as limitations of the metrics? Seems odd that we point out shortcomings of the metrics and go on trying to build approaches for these metrics.}
%However, these two criteria exist in conflict with one another, and a significant body of work \cite{locatello2019fairness, golz2019paradoxes,corbett2018measure,barocas-hardt-narayanan} has gone into explaining that independence and separation are inherently incompatible for non-trivial cases. Moreover, each criteria is capable of harming the very groups it seeks to protect if applied to the wrong real-world problem, which motivates us to construct a framework that is flexible enough to handle different fairness criteria.

%Furthermore, some works \cite{sharifi2019average} also look into different ways to formulate fairness which may be more relevant to a specific use case.

Different contexts also require different points of intervention during the learning process to ensure fairness. \textit{Pre-processing}  \cite{KamiranCalders2012,zemel2013learning,FeldmanFMSV2015,calmon2017optimized, sattigeri2018fairness, xu2018fairgan} approaches modify the data to eliminate bias whereas  \textit{post-processing}  \cite{Kamiran2012, hardt2016equality, pleiss2017fairness,wei2019optimized} modify learned features/predictions from existing models to be more fair. We focus on the \textit{in-processing} approach \cite{kamishima2012fairness,zhang2018mitigating,celis2019classification, mary2019fairness}, where the fairness criteria are directly incorporated in the training objective to produce fairer learned features. Motivated by few-shot applications where only a pre-trained network and few samples labeled with the sensitive attribute are available, we also want our methods to be applicable in a post-processing manner. 

%In this few-shot regime, by applying the in-processing training algorithms to a pre-trained network,  fairness can still be enforced after-the-fact.

In this paper, we frame the ideas of independence and separation in a way which allows a relevant regularizer or penalty term to be derived in addition to a measure of fairness, that is useful in enforcing fairness while also tractable, admitting an optimization algorithm (e.g. if used as an objective for a neural net trained using gradient descent, it must be differentiable), and easily computed. Existing approaches can struggle with efficiency, can fail to provide good control over the performance-fairness tradeoff, and/or can only deal with either discrete or continuous data.

We make the following contributions in this paper:

\begin{itemize}
    \item We present a universal framework justified by an information-theoretic view that can inherently handle the popular fairness criteria, namely independence and separation, while seamlessly adopting both discrete and continuous cases, which uses the maximal correlation to construct measures of fairness associated with different criteria, then uses these measures to further develop fair learning algorithms in a fast, efficient and effective manner.
    \item We show empirically that these algorithms can provide the desired smooth tradeoff curve between the performance and the measures of fairness on several standard datasets (COMPAS, Adult, and Communities and Crimes), so that a desired level of fairness can be achieved.
    \item Finally, we perform experiments to illustrate that our algorithms can be used to impose fairness on a model originally trained without any fairness constraint in the few-shot regime, which further demonstrates the versatility of our algorithms in a post-processing setup.
\end{itemize}

\section{Background}
\subsection{Fairness Objectives in Machine Learning}

Consider the standard supervised learning scenario where we predict the value of a target variable $Y\in\cY$ using a set of decision or predictive variables $X\in\cX$ with training samples $\{(x_1,y_1),\dots,(x_n,y_n)\}$. For example, $X$ may be information about an individual's credit history, and $Y$ is whether the individual will pay back a certain loan. In general, we wish to find features $f(x)$, which are predictive of $Y$, so that we can construct a good predictor $\hat{y}=T(f(x))$ of $y$ under some loss criteria $L(\hat{y},y)$.

Now suppose we have some sensitive attributes $D\in\cD$ we wish to be ``fair'' about (e.g. race, gender), and training samples $\{(x_1,y_1,d_1),\dots,(x_n,y_n,d_n)\}$. For example, in the criminal justice system, predictions about the chance of recidivism of a convicted criminal ($Y$) given factors such as the nature of the crime and the number of prior arrests ($X$) should not be determined by race ($D$). This is a known issue with the COMPAS recidivism score, which, despite not using race as an input to make decisions, still leads to systematic bias towards members of certain races in the output score \cite{angwin2016machine,chouldechova2017fair}.

The two most popular criteria for fairness are independence and separation. Independence states that for a feature to be fair, it must satisfy the independence property $\hat{Y} \perp D$ or $R \perp D$, where $R$ is a ``score'' function (equivalent to $f(x)$ in this paper). The intuition is simple: if the score/prediction is independent of the sensitive attribute, then no information about the sensitive attribute is used to predict $Y$. This criterion has been studied under the lens of \textit{demographic parity} and \textit{disparate impact} \cite{barocas-hardt-narayanan}, and admits a class of fairness measures based on the degree of dependence between $f(X)$ and $D$. For example, independence is satisfied if and only if the mutual information $I(f(X);D)$ is zero. When $D$ is binary, another popular class of measures can be defined by $\mathbb{D}(\pr(Y|D=1);\pr(Y|D=0))$ (For example, the US Equal Employment Opportunity Commission \cite{equal1978department} uses disparate impact $\mathbb{D}(\pr(Y|D=1);\pr(Y|D=0)) = \frac{\pr(\hat{Y}=1|D=0)}{\pr(\hat{Y}=1|D=1)}$ ).

Separation requires the conditional independence property $(\hat{Y} \perp D)|Y$ or  $(f(X) \perp D)|Y$. This criterion allows for violation of demographic parity to the extent that it is justified by the target variable. In the general case, this criterion suggests a fairness measure based on the conditional dependence between $\hat{Y}$ and $D$ conditioned on $Y$. In the case where $D$ is binary, we obtain the \textit{equalized opportunities} (EO) measures \cite{barocas-hardt-narayanan}, which are given by the differences in error rates for the two groups (e.g. the difference between the false positive rates for $D=0,1$). For a more complete discussion of the advantages and disadvantages of these two criteria, please refer to \cite{barocas-hardt-narayanan}.

%Equipped with these two criteria and their associated measures, we seek to find ways to enforce them in a tractable and efficient manner.

\subsection{Maximal Correlation}

Since these fairness criteria are expressed as enforcing independencies with respect to joint distributions, we look for constraints that reduce dependency between variables. In particular, the right formulation of correlation between learned features and sensitive attributes can provide a framework for measuring and optimizing for fairness. One effective measure applicable to both continuous and discrete data is the Hirschfeld-Gebelein-Renyi (HGR) maximal correlation, a measure of nonlinear correlation which originated in \cite{hoh35} and is further developed in \cite{hg41,ar59}. The HGR maximal correlation between two random variables is equal to zero if and only if the two variables are independent, and increases in value the more correlated they are (i.e., the more biased/unfair).

\begin{definition}
For two jointly distributed random variables $X \in \cX$ and $Y \in \cY$, given $1\le k\le K-1$ with $K=\min\{|\cX|,|\cY|\}$,
the HGR maximal correlation problem is
\begin{equation}
\label{hgr}
(\bff^*,\bg^*) \defeq \hspace{-0.5cm} \argmax_{\substack{\bff \colon \cX \to
    \reals^k,\ \bg \colon \cY 
    \to
    \reals^k} }\hspace{-0.5cm}
\E{\bff^\T(X)\,\bg(Y)},
\end{equation}
with $\E{\bff(X)}=\E{\bg(Y)}=\bzero$,  $\E{\bff(X)\bff^\T(X)}=\E{\bg(Y)\bg^\T(Y)}=\bI$, and expectations taken over $P_{X,Y}$.
We refer to $\bff^*$ and $\bg^*$ as maximal correlation
functions,  with $\bff^* = (f^*_1,\dots,f^*_k)^\T$ and $\bg =
(g^*_1,\dots,g^*_k)^\T$, and the associated maximal correlations are
\begin{equation}
    \sigma(f^*_i\,g^*_i) \defeq \E{f^*_i(X)\,g^*_i(Y)},\ \text{for }\ i=1,\dots,k,
\end{equation}
and the HGR maximal correlation is
\begin{equation}
\HGR_k(X,Y) \defeq \E{\bff^{*\T}(X)\,\bg^*(Y)} =\sum_{i=1}^k \sigma(f^*_i\,g^*_i).
\end{equation}
\end{definition}

Note that the original definition of HGR maximal correlation is the special case of our definition when $k=1$ \cite{hmwz19}. This generalization of maximal correlation analysis enables us to produce more than one feature mapping by solving the maximal correlation problem, and these feature mappings can be used in other applications, including ensemble learning, multi-task learning and transfer learning \cite{lee2019learning,wang2019efficient}.

%maximal correlation analysis originated with the work of Hirschfeld \cite{hoh35}, and has
%been further developed in subsequent works by Gebelein and R\'enyi \cite{hg41,ar59}. Thus, it is often referred to as HGR maximal correlation analysis. %(For a more detailed summary of this literature, see, e.g., the references and discussion in .)

\subsection{Related Work}

Independence and separation have been studied in many works. Most existing approaches fail to provide an efficient solution in both discrete/continuous settings. \cite{zemel2013learning} develops an optimizer using absolute difference in odds $|\pr(\hat{Y}=1|D=1)-\pr(\hat{Y}=1|D=0)|$ as a regularizer, which requires discrete $Y$ and $D$ and was only applied to Na\"{i}ve Bayes and Logistic Regression to enforce the independence criterion. %They also use the absolute difference in odds as the measure of fairness. 
In \cite{hardt2016equality}, a post-processing method is provided using probabilistic combination of classifiers to achieve desired ROC curves, which only applies when $D$ is discrete.
Alternatively, \cite{calmon2017optimized} propose pre-processing the data beforehand to enforce fairness before learning, based on randomized mappings of the data subject to a fairness constraint defined by $J = \max(|\frac{\pr(\hat{Y}=1|D=1)}{\pr(\hat{Y}=1|D=0)} - 1|,|\frac{\pr(\hat{Y}=1|D=0)}{\pr(\hat{Y}=1|D=1)} - 1|)$. Again, this method is only designed for independence with discrete $Y$ and $D$, and requires processing the entire dataset, which is computationally complex.

Other methods can also be limited in their ability to handle all dependencies between variables. \cite{zafar2017fairness} uses a covariance-based constraint to enforce fairness, so it likely would not do well on other metrics. Furthermore, it is strictly a linear penalty rather than our non-linear formulation and penalizes the predictions of the system rather than the features learned. This limits the relationships between variables it can capture. An adversarial method is proposed in \cite{zhang2018mitigating} to enforce independence or separation, but requires the training of an adversary to predict the sensitive attribute, which can introduce issues of convergence and bias.

%result in slower performance.

Recently, \cite{mary2019fairness} propose the use of the HGR maximal correlation %as an approximation for mutual information to act 
as a regularizer for either the independence or the separation constraint. In contrast to our approach dealing with the maximal correlation directly, they use a $\chi^2$ divergence  computed %with a discretization of these variables 
over a mesh grid to upper bound the HGR maximal correlation %for the purposes of computing the gradient 
during the optimization of the classifier (either a linear regressor or a Deep Neural Net (DNN)). This method applies to cases where $X$ is continuous and $Y$ and $D$ are either continuous or discrete variables, but scales poorly with the bandwidth and dimensionality of $D$, and treats the discrete case in the same way as the continuous case, resulting in slow performance on discrete datasets. %\prasanna{Can we briefly say how our method improves on this or just mention the drawbacks.}

There are other works which use either an HGR-based or mutual information-based formulation of fairness, but do not generalize to more than one setting. \cite{grari2019fairness} and \cite{baharlouei2019r} use correlation-based regularizers, but can only be used in the independence case. Furthermore, \cite{baharlouei2019r} only uses a single mode of the HGR maximal correlation (as opposed to multiple modes, which our method will use) for regularization, which limits the information it can encapsulate, and is also not designed for continuous sensitive attributes. \cite{moyer2018invariant} also develops a method which can only be used for independence, and requires training an additional network in order to evaluate a bound for the mutual information which can be used to as a fairness penalty, thus increasing the complexity and required runtime.
Finally, \cite{cho2020fair} approximates the mutual information with a variational approximation of the mutual information, but does not include a formulation for continuous labels.

%\cite{mary2019fairness} propose the use of the HGR maximal correlation as an approximation for mutual information to act as a regularizer when training a classifier with either the independence or the separation constraint, then uses a $\chi^2$ divergence between $f(X)$ and $D$ computed with a discretization of these variables over a mesh grid to upper bound the HGR maximal correlation for the purposes of computing the gradient during the optimization of the classifier (either a linear regressor or a Deep Neural Net (DNN)). This method can be applied to both continuous and discrete variables. Similarly, \cite{grari2019fairness} uses an approximation of the HGR as a regularizer in which a pair of zero-mean unit-variance functions are trained to optimize the maximal correlation between $f(X)$ and $D$ for the continuous case (as opposed to our method, which will make use of the variance-penalized Soft-HGR instead), though their method would likely work with discrete variables as well. However, their formulation only works for independence and \textit{Equalized Residuals}, rather than separation/equalized opportunities. Additionally, they do not provide a method for extending to separation, due to the lack of a definition for the conditional HGR.

\section{Maximal Correlation for Fairness}

Equipped with the HGR maximal correlation as a measure of dependence, we explore its use as a fairness penalty. Depending on the data modality (discrete/continuous) and the fairness criteria (independence/separation), the resulting fair learning algorithm takes different specifically-tailored forms. In this section, we demonstrate how to derive these regularizers and algorithms to ensure the aforementioned fairness objectives for both discrete and continuous cases.

% \begin{equation}\label{equ:hgr}
% (\bff^*,\bg^*) \defeq \argmax_{\substack{\bff \colon \cX \to
%     \reals^k,\ \bg \colon \cY 
%     \to
%     \reals^k\\ \E{\bff(X)}=\E{\bg(Y)}=\bzero,
%     \\ \E{\bff(X)\bff^\T(X)}=\E{\bg(Y)\bg^\T(Y)}=\bI}}
% \E{\bff^\T(X)\,\bg(Y)},
% \end{equation}
%[Definition and overview of HGR as approximation for MI]
\subsection{Maximal Correlation for Discrete Learning}

In this subsection, the decision variable $X$, target variable $Y$, and sensitive attribute $D$ are discrete random variables defined on alphabets $\cX$, $\cY$ and $\mathcal{D}$, respectively. We first describe how to solve the discrete maximal correlation problem using a Divergence Transfer matrix (DTM)-based approach.

\begin{definition}
The $(y,x)$\/th entry of the divergence transfer matrix $\bB_{Y,X}\in \reals^{|\cY|\times|\cX|}$ associated with joint distribution $P_{X,Y}$ is given by:
$\bB_{Y,X}(y,x) \triangleq \frac{P_{X,Y}(x,y)}{\sqrt{P_X(x)}\sqrt{P_Y(y)}}.$
\end{definition}

The following useful result (see, e.g., \cite{hmwz19}) expresses that the maximal correlation problem can be  solved by simply computing the singular value decomposition (SVD) of $\bB$ in the discrete case.

\begin{theorem}\label{thm:main}
Assume that the SVD of DTM $\bB_{Y,X}$ takes the form
% \begin{equation}\label{eq:SVD}
$\bB_{Y,X} = \sum_{i=0}^{K-1} \sigma_i \psi_i^Y (\psi_i^X)^\T$,
% \end{equation}
with singular values $\sigma_0\ge\sigma_1\ge \dots\ge \sigma_{K-1}$, singular vectors $\psi_i^Y$, $\psi_i^X$, and $K=\min\{|\cX|,|\cY|\}$. Then we have $\sigma_0=1,\ \psi_0^X(x)=\sqrt{P_X(x)}, \ \psi_0^Y(y)=\sqrt{P_Y(y)}$,
and the maximal correlation functions are related to the singular vectors in the SVD:
$f_i^*(x)=\frac{\psi_i^X(x)}{\sqrt{P_X(x)}},\  g_i^*(x)=\frac{\psi_i^Y(y)}{\sqrt{P_Y(y)}}$, 
with associated maximal correlations
$\sigma(f^*_i\,g^*_i) = \sigma_i$, for $i=1,\cdots,K-1$. Thus, the conditional distribution $P_{Y|X}$ has the following decomposition:
\begin{equation}\label{eq:cond_dis}
P_{Y|X}(y|x)=P_{Y}(y)\Big[1+\sum_{i=1}^{K-1}\sigma_if^*_i(x)g^*_i(y
    ) \Big].
\end{equation}
\end{theorem}
As we can see from this theorem, the singular values $\sigma_i$ \footnote{Since the associated maximal correlations equals to the corresponding singular values of DTM, we abuse the notation a little bit and use $\sigma$ to denote both of them.} of the matrix $\bB_{Y,X}$ essentially characterizes the dependence between two discrete random variables, and the singular vectors $\Phi^X = [\psi_1^X,\cdots, \psi_k^X ]$ and $\Phi^Y = [\psi_1^Y,\cdots, \psi_k^Y ]$ are equivalent to the maximal correlation functions $\bff$ and $\bg$.

Since our goal is to construct feature mappings $\bff(x)$ under fairness constraints, we will use the following variational characterization of an SVD, which does not involve $\bg(y)$:
\begin{lemma}\label{lemma:svd} \cite{horn2012matrix}
For any $k\le K-1$ and $\Phi_X \in \mathbb{R}^{|\cX|\times {(k+1)}}$, $\max_{\Phi_X^\T \Phi_X=\bI} \|\bB\Phi_X\|_{\mathrm{F}}^2=\sum_{i=0}^{k} \sigma^2_i$, where  $\|A\|_{\mathrm{F}} \defeq \sqrt{\tr(A^\T A)}$ denotes the Frobenius norm of matrix $A$.
\end{lemma}

\subsubsection{Independence:}
To ensure sufficient independence, we must construct feature mappings $\bff: \cX\to \reals^{k} $ so that the maximal correlations between $\bff(X)$ and $Y$ are large, while the ones between $\bff(X)$ and $D$ are small. Thus, we propose the following DTM-based approach to construct $\bff$:
\begin{equation}\label{eq:indep_loss}
    \max_{\Phi \in \mathbb{R}^{|\cX|\times (k+1)}: \Phi^\T \Phi=\bI} \|\bB_{Y,X} \Phi \|_{\mathrm{F}}^2 -\lambda \|\bB_{D,X} \Phi \|_{\mathrm{F}}^2,
\end{equation}
where $\bB_{Y,X}$, $\bB_{D,X}$ denote the DTMs of distribution $P_{Y,X}$ and $P_{D,X}$, respectively, and $\lambda$ is the regularization coefficient that controls the penalty of the maximal correlations between $\bff(X)$ and $D$. $\Phi^*=[\phi_0^*,\phi_1^*,\cdots, \phi_k^* ]$ is the solution of the optimization problem \eqref{eq:indep_loss}. As shown in Theorem \ref{thm:main}, $\bB_{Y,X}$, $\bB_{D,X}$ have a shared right singular vector $\sqrt{P_X(x)}$, and we can let $\phi_0^* = \sqrt{P_X(x)}$. Then, the feature mappings for independence can be obtained by normalizing other column vectors in $\Phi^*$
\begin{equation}
    f_i(x) = \phi_i^*(x)/\sqrt{P_X(x)},\ i=1,\cdots,k.
\end{equation}

%where 
%$\bff(x) = F^*(x)/\sqrt{P_X(x)}$.   

We have the following remarks:

1) The optimization problem in \eqref{eq:indep_loss} can be written as $\max \tr(\Phi^\T \big( \bB_{Y,X}^\T \bB_{Y,X} - \lambda \bB_{D,X}^\T \bB_{D,X}\big)\Phi)$, and can be solved exactly by computing the eigen-decomposition of $\bB_{Y,X}^\T \bB_{Y,X} - \lambda \bB_{D,X}^\T \bB_{D,X}$. Thus, the proposed approach is more computationally efficient than existing gradient-descent algorithms.

2) Lemma \ref{lemma:svd} states that the Frobenius norm squared $\|\bB_{Y,X} F \|_{\mathrm{F}}^2$ corresponds to the squared sum of the singular values. Actually, the following lemma shows that $\|\bB_{Y,X} F \|_{\mathrm{F}}^2$ can be further related to the mutual information $I(X;Y)$ when the dependence between $X$ and $Y$ is weak.
    \begin{lemma}\cite{hmwz19}\label{lemma:MI}
Let $X\in \cX$ and $Y\in \cY$ be $\epsilon$-dependent random variables, i.e., the $\chi^2$-divergence is bounded
$D_{\chi^2}(P_{X,Y}\| P_X P_Y)\le \epsilon$, then $I(X;Y)=\frac{1}{2} \sum_{i=1}^{K-1} \sigma_i^2 +o(\epsilon^2)$.
\end{lemma}

3) %Since we want to minimize the dependence between $\bff(X)$ and $D$, which implies the weak dependence, 
As suggested by Lemma \ref{lemma:MI}, the optimization problem in \eqref{eq:indep_loss} is approximately maximizing the mutual information between  $\bff(X)$ and $Y$, while penalizing the mutual information $I(\bff(X);D)$.

Once we  solve  \eqref{eq:indep_loss} and obtain the feature mappings $\bff(x)$, we can obtain the corresponding maximal correlation function $\bg(y)$ for the target variable $Y$ via one step of the alternating conditional expectations algorithm by \cite{bf85}: ${g}_i(y) \propto \Ed{p_{X|Y}(\cdot|y)}{f_i(X)}, \  i=1,\dots,k$. In turn, $\bg(y)$ can be computed by further normalizing the conditional expectations of $\bff(X)$, so that the condition $\E{\bg(Y)\bg^\T(Y)}=\bI$ is satisfied. Finally, the predictions $\hat{Y}$ can be made following the Maximum A Posteriori (MAP) rule, where the posteriori distribution $P_{{Y}|X}(y|x)$ can be approximately computed  by 
plugging the learned feature mappings $\bff(X)$ and $\bg(Y)$ into \eqref{eq:cond_dis}, i.e.,
\begin{equation}\label{eq:map}
\hat{Y} %= \argmax_{y\in \cY} \hat{P}_{Y|X'}(y|x') \nonumber \\
     =\argmax_{y\in \cY} {P}_{Y}(y)\Big[1+\sum_{i=1}^{k}{\sigma}_i {f}_i(x){g}_i(y
    ) \Big].
\end{equation}

\subsubsection{Separation:}
For the separation criterion, we want to ensure sufficient conditional independence $(f(X) \perp D)|Y$. Here, we cannot simply replace the $\bB_{D,X}$ in \eqref{eq:indep_loss} with a conditional DTM, as it involves three random variables and thus cannot be usefully expressed as a matrix. Since maximal correlation is related to mutual information as shown in Lemma \ref{lemma:MI}, we consider the following formulation: %we construct $\bff$ by maximizing $I(\bff(X);Y)$, while penalizing the joint mutual information $I(\bff(X);D,Y)$. Then
\begin{align}\label{eq:sep_loss_MI}
    &\max_{\bff} I(\bff(X);Y)-\lambda I(\bff(X);D,Y)\nonumber\\
    &= \max_{\bff} I(\bff(X);Y)-\lambda \big(I(\bff(X);Y)+ I(\bff(X);D|Y)\big)\nonumber\\
    & = \max_{\bff} (1-\lambda)I(\bff(X);Y)-\lambda  I(\bff(X);D|Y),
\end{align}

where the first equality follows from the chain rule of mutual information and a given $\lambda\in(0,1)$. Thus, we can control the conditional mutual information $I(\bff(X);D|Y)$, by adding the joint mutual information $I(\bff(X);D,Y)$ as a regularizer in the training process.

Note that Lemma \ref{lemma:svd} and Lemma \ref{lemma:MI} implies that mutual information can be approximated using DTM as shown in \eqref{eq:indep_loss} in independence case. Accordingly, we approximate \eqref{eq:sep_loss_MI} using the following optimization problem  to ensure the separation criterion for discrete data:
\begin{equation}\label{eq:sep_loss}
    \max_{\Phi \in \mathbb{R}^{|\cX|\times (k+1)}: \Phi^\T \Phi=\bI} \|B_{Y,X} \Phi \|_{\mathrm{F}}^2 -\lambda \|B_{D\otimes Y,X} \Phi \|_{\mathrm{F}}^2,
\end{equation}
where $D\otimes Y$ is the Cartesian product of $D$ and $Y$, and $B_{D\otimes Y,X}$ denotes the DTM of distribution $P_{D\otimes Y,X}$. Once we obtained the solution $\Phi^*$, we could follow similar steps as in the independence case to get $\bff(x)$ and $\bg(y)$, and make predictions for the test samples.

% \begin{align}
%     &I(X;Y)-\lambda I(X;D)\\
%     & = \frac{1}{2} \sum_{i=1}^{K-1} \sigma_{i,XY}^2 - \frac{1}{2} \sum_{i=1}^{K-1} \sigma_{i,XD}^2\\
%     & = \max_{\{F\in \mathbb{R}^{k_2\times k}: F^\T F=\bI\}}\|AF\|_{\mathrm{F}}^2 - \max_{\{F\in \mathbb{R}^{k_2\times k}: F^\T F=\bI\}}\|AF\|_{\mathrm{F}}^2\\
%     &\le \max_{\{F\in \mathbb{R}^{k_2\times k}: F^\T F=\bI\}}\|B_{X,Y}F\|_{\mathrm{F}}^2 - \|B_{X,D}F\|_{\mathrm{F}}^2
% \end{align}

% The loss function for discrete random variables for the Independence criterion:
% \begin{equation}
%     \max_{F\in \mathbb{R}^{K\times k}: F^\T F\bI} \|B_{X,Y} F \|_{\mathrm{F}}^2 -\lambda \|B_{X,D} F \|_{\mathrm{F}}^2
% \end{equation}

\subsection{Maximal Correlation for Continuous Learning}

When $X$, $Y$, and $D$ are all continuous and real-valued, computing the HGR maximal correlation becomes much more difficult since the space of functions over real numbers is not tractable. We thus turn to approximations, and begin by limiting our scope of learning algorithms to those which train models (e.g. neural nets) via gradient descent (or SGD) using samples, which encompasses most commonly-used methods. 
It then follows that any approximation of the HGR maximal correlation used must be differentiable to calculate the gradient. We thus restrict the space of maximal correlation functions to be the family of functions that can be learned by neural nets, allowing us to compute the gradient while still providing a rich set of functions to search over.

\subsubsection{Independence:} 
To ensure sufficient independence, we want to minimize the loss function $L(\hat{Y},Y)$ and the maximal correlation between $\bff(X)$ and $D$. Our optimization (for a given $\lambda$) then becomes:
\begin{equation}
    \min_{\substack{\bff \colon \cX \to
    \reals^m\\T \colon \reals^m \to \cY}} L(T(\bff(X)),Y) + \lambda\HGR_k(\bff(X),D),
\end{equation}
where   $\HGR_k(\bff(X),D) = \max_{\substack{\bg,\ \bh }}
 \E{\bg^\T(\bff(X))\,\bh(D)}$, with     $\E{\bg(\bff(X))}=\E{\bh(D)}=\bzero$, and
$\E{\bg(\bff(X))\bg^\T(\bff(X))}=\E{\bh(D)\bh^\T(D)}=\bI$.
% where 
% \begin{equation}
%     C %&= L(T(\bff(X)),Y) + \lambda\HGR_k(\bff(X),D) \\ \nonumber
%     = L(T(\bff(X)),Y) + \lambda\E{\bg^\T(\bff(X))\,\bh(D)}.
% \end{equation}
$m$ is the dimension of the features $\bff(X)$, $k$ is the number of maximal correlation functions, and $\bg \colon \reals^m \to \reals^k,\ \bh \colon \cD \to \reals^k$ are the maximal correlation functions relating $\bff(X)$ with $D$. Given the difficulty of enforcing the orthogonalization constraint, we use a variational characterization of the HGR maximal correlation called Soft-HGR \cite{wang2019efficient} which relaxes this constraint:
\begin{align*}
&\HGR_{\soft}(X,Y) \defeq \\ \hspace{-0.5cm} &\max_{\substack{%\bff \colon \cX \to     \reals^k\\ \bg \colon \cY     \to\reals^k\\ 
    \E{\bff(X)}=\bzero\\\E{\bg(Y)}=\bzero
    }} 
\!\E{\bff^\T(X)\,\bg(Y)}\!- \frac{1}{2}\tr(\cov(\bff(X))\cov(\bg(Y))),
\end{align*}
where $\cov(X)$ is the covariance matrix of $X$. Then, our learning objective becomes:
\begin{equation}\label{eq:cts_ind_loss}
     \min_{\substack{\bff \colon \cX \to
    \reals^m\\T \colon \reals^m \to \cY}}   \max_{\substack{\bg \colon \reals^m \to
    \reals^k,\ \bh \colon \cD
    \to
    \reals^k\\ \E{\bg(\bff(X))}=\E{\bh(D)}=\bzero}}   C,
\end{equation}
where $C=L(T(\bff(X)),Y)  + \lambda\Big(\!\E{\bg^\T(\bff(X))\,\bh(D)}\!\!-\!\! \frac{1}{2}\!\tr\!\big(\!\!\cov(\bg(\bff(X)))\cov(\bh(D))\big)\Big)$.
     
We solve this optimization by alternating between optimizing $\bff,T$ and optimizing $\bg,\bh$. In practice, we implement this by alternating between one step of gradient descent for $\bff$ and $T$ and 5 steps of gradient descent on $\bg$ and $\bh$ to allow the maximal correlation functions to adapt to the changing $\bff$.

\subsubsection{Separation:} 
For separation, we use a similar argument as in the discrete case to ensure the conditional independence. Specifically, we solve the following optimization problem:
\begin{align} \label{eq:cts_sep_loss}
    % \max_{\substack{\bg \colon \reals^m \to
    % \reals^k,\ \bh \colon \cY\times\cD 
    % \to
    % \reals^k\\ 
    % \bg' \colon \reals^m \to
    % \reals^k,\ \bh' \colon \cD
    % \to
    % \reals^k\\ 
    % \E{\bg(X)}=\E{\bh(Y)}=\bzero\\
    % \E{\bg'(X)}=\E{\bh'(Y)}=\bzero}}
    &\min_{\substack{\bff \colon \cX \to
    \reals^m\\T \colon \reals^m \to \cY}}  L(T(\bff(X)),Y) +  \\\nonumber
    &\hspace{0.3in}\lambda\big(\HGR_{\soft}(f(X),D\otimes Y) - \HGR_{\soft}(f(X),Y)\big).
\end{align}
% Where
% \begin{multline}
%         C = L(T(\bff(X)),Y) \\ + \lambda(\HGR_{\soft}(f(X),(D,Y)) - \HGR_{\soft}(f(X),D))
% \end{multline}
Note that for the first soft-HGR term, we use $\bg,\bh$ to denote the maximal correlation functions, and $\bg',\bh'$ to denote the functions for the second term. Similar to the discrete case, the difference term allows us to approximate the conditional mutual information using two unconditional terms. Once again, we solve this optimization by alternating between optimizing $\bff,\ T$ and optimizing $\bg,\bh,\bg',\bh'$.

\section{Experimental Results}

%\prasanna{We can create tables for fig1-6 and put in appendix if the readers want to see actual number.}\joshua{sure}
In order to illustrate the effectiveness of our algorithms, we run experiments using the proposed algorithms on discrete (Adult and COMPAS) and continuous (Communities and Crimes) datasets.

\subsection{Discrete Case}
%\prasanna{We should say why we dont compare with Mary et al in the discrete case. If time permit can we try running the chi-sq code with only discrete features?}\joshua{updated related works to show that Mary's method is only for continuous X}
We test the proposed DTM-based approach on the ProPublica’s COMPAS recidivism dataset\footnote{https://github.com/propublica/compas-analysis} and the UCI Adult dataset\footnote{https://archive.ics.uci.edu/ml/datasets/adult}, which were chosen as they contain categorical features and are used in prior works. More experiments for the discrete case can be found in the supplementary materials. 

For the COMPAS dataset, the goal is to predict whether the individual recidivated (re-offended) ($Y$) using the severity of charge, number of prior crimes, and age category as the decision variables ($X$). As discussed in \cite{calmon2017optimized}, COMPAS scores are biased against African-Americans, so race is set to be the sensitive attribute ($D$) and filtered to contain only Caucasian and African-American individuals. %(The encoding of categorical variables is described in the SM.) 
As for the Adult dataset, the goal is to predict the binary indicator ($Y$) of whether the income of the individual is more than 50K or not based on the following decision variables ($X$):  age (quantized to decades) and education (in years), and the sensitive attribute ($D$) is the gender of the individual.

For both datasets, we randomly split all data into 80\%/20\% training/test samples.
We first construct an estimate of DTM $\hat{\bB}$   with the empirical distribution of the training set, then solve the proposed optimization in \eqref{eq:indep_loss} and \eqref{eq:sep_loss} using $\hat{\bB}$ to obtain fair feature mappings $\hat{\bff}(x),\hat{\bg}(y)$. The predictions $\hat{Y}$ of the test samples $X'$ are given by plugging the learned feature mappings $\hat{\bff}(x'),\hat{\bg}(y)$ into the MAP rule \eqref{eq:map}, where ${P}_{Y}$ can be estimated from the empirical distribution $\hat{P}_{Y}$ on the training set.
% \begin{align}
%     \hat{Y} %&= \argmax_{y\in \cY} \hat{P}_{Y|X'}(y|x') \nonumber \\
%     & =\argmax_{y\in \cY} \hat{P}_{Y}(y)\Big[1+\sum_{i=1}^{K-1}\hat{\sigma}_i \hat{f}_i(x')\hat{g}_i(y
%     ) \Big],
% \end{align}
%

\begin{figure}[t]
\centering
\begin{minipage}{\linewidth}
        \includegraphics[width=\linewidth]{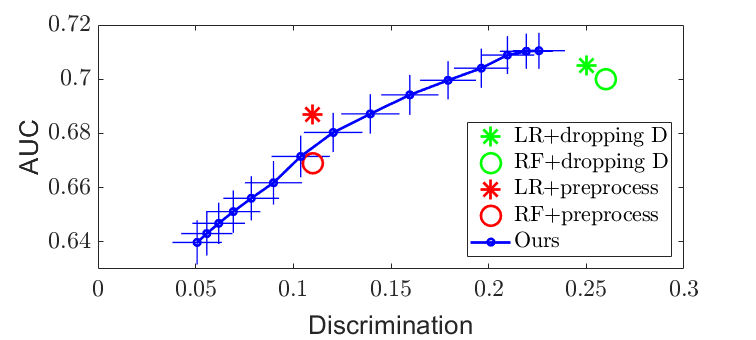}
        \centering
        \end{minipage}\\
\begin{minipage}{\linewidth}
        \includegraphics[width=\linewidth]{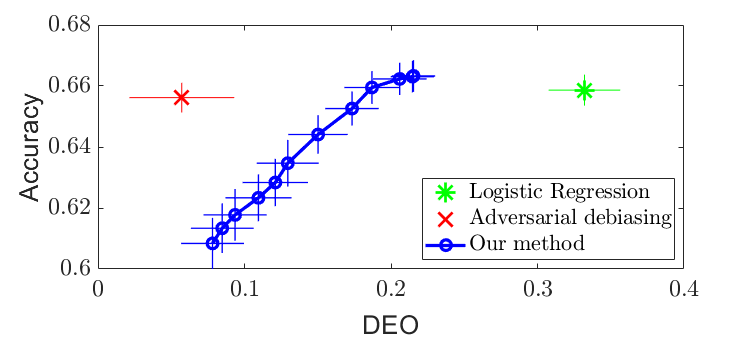}
        \centering
        \end{minipage}
\caption{Regularization results on COMPAS dataset, with AUC plotted against discrimination measure for independence (Top), and accuracy plotted against DEO for separation (Bottom), respectively.}\label{fig:discrete_compas}
\end{figure}

\begin{figure}[t]
\centering
\begin{minipage}{\linewidth}
        \includegraphics[width=\linewidth]{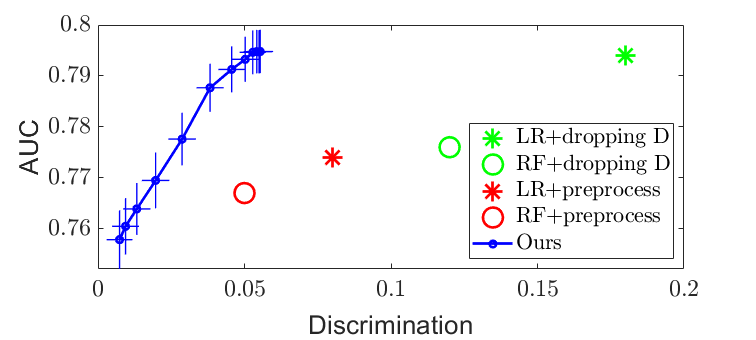}
        \centering
        \end{minipage}\\
\begin{minipage}{\linewidth}
        \includegraphics[width=\linewidth]{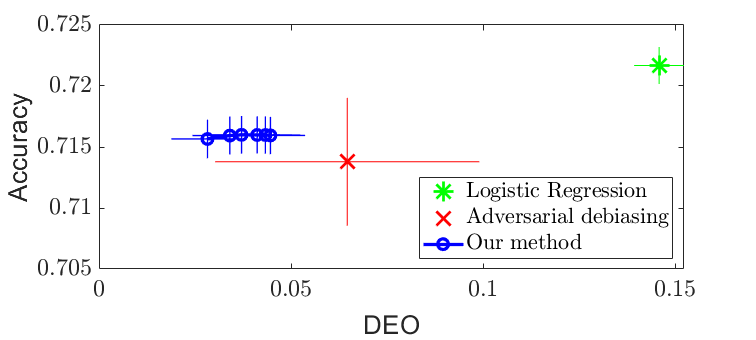}
        \centering
        \end{minipage}
\caption{Regularization results on Adult dataset, with AUC plotted against discrimination measure for independence (Top), and accuracy plotted against DEO for separation (Bottom), respectively.}\label{fig:discrete_adult}
\end{figure}

For the independence case, we compare the trade-off between the performance and the discrimination achieved by our method with that of the optimized pre-processing methods proposed in \cite{calmon2017optimized}. Note that we adopt the same settings as the experiments in \cite{calmon2017optimized} to do a fair comparison, and the reported results for their method are from their work. We plot the area under ROC curve (AUC) of $\hat{P}_{Y|X'}(y|x')$ compared to the true test labels $Y'$ against the following standard discrimination measure derived from legal proceedings \cite{equal1978department}:
% \begin{equation}\label{eq:J_measure}
$J=\max_{d,d'\in\mathcal{D}}\big|\mathbb{P}_{\hat{Y}|D}(1|d)/\mathbb{P}_{\hat{Y}|D}(1|d')-1 \big|$.
% \end{equation}
Figures \ref{fig:discrete_compas} and \ref{fig:discrete_adult} (Top)  shows the results. For both datasets, it can be seen that simply dropping the sensitive attribute $D$ and applying logistic regression (LR) and random forest (RF) algorithms cannot ensure independence between $\hat{Y}$ and $D$. However, the proposed DTM-based algorithm provides a trade-off between performance and discrimination by varying the value of the regularizer $\lambda$ in the optimization \eqref{eq:indep_loss}, which outperforms the optimized pre-processing methods in \cite{calmon2017optimized} on the Adult dataset, and achieves similar performance on the COMPAS dataset. More importantly, the DTM-based algorithm provides a smooth trade-off curve between the performance and discrimination, so that a desired level of fairness can be achieved by setting $\lambda$ in practice. In addition, since our method only requires us to perform eigen-decomposition, it runs significantly faster than the optimized pre-processing method, which needs to solve a much more complex optimization problem. Empirically, we find at least a tenfold speed up in runtime compared to the existing methods.

For the separation criterion, we compare the balanced-accuracy achieved by our algorithm with that of the adversarial debiasing method in \cite{zhang2018mitigating} against the difference in equalized opportunities (DEO), another standard measure used commonly in the literature: $\mathrm{DEO}=\pr(\hat{Y}\!\!=\!\!1|D\!\!=\!\!1,Y\!\!=\!\!1)\!-\!\pr(\hat{Y}\!\!=\!\!1|D\!\!=\!\!0,Y\!\!=\!\!1)$.
The results on the COMPAS and Adult datasets are presented in Figures \ref{fig:discrete_compas} and \ref{fig:discrete_adult} (Bottom). Compared to the na\"{i}ve logistic regression, the proposed DTM-based algorithm dramatically decreases the DEO while maintaining similar accuracy performance on both datasets, which outperforms the adversarial debiasing method in \cite{zhang2018mitigating} on the Adult dataset. %Since the adversarial debiasing method used all the available decision variables to make decisions, their accuracy on COMPAS  is better than that of the proposed algorithm.  
We note that the accuracy and DEO curve achieved by the proposed algorithm in the separation setting has a smaller range compared to that in the independence setting. This is because the value of the regularizer $\lambda$ is restricted in the separation optimization problem \eqref{eq:sep_loss} to $\lambda\in [0,1)$, but only to $\lambda>0$ for the optimization problem \eqref{eq:indep_loss}.

% \begin{figure}[ht]
% \vskip 0.2in
% \begin{center}
% \centerline{\includegraphics[width=\columnwidth]{Compas_Separation_new.png}}
% \caption{Results for Separation regularization on the discrete COMPAS dataset, accuracy is plotted with respect to the DEO.}
% \label{COMPAS_sep}
% \end{center}
% \vskip -0.2in
% \end{figure}

% \begin{figure}[ht]
% \vskip 0.2in
% \begin{center}
% \centerline{\includegraphics[width=\columnwidth]{Adult_Separation_new.png}}
% \caption{Results for Separation regularization on the discrete Adult dataset, accuracy is plotted with respect to the DEO.}
% \label{Adult_sep}
% \end{center}
% \vskip -0.2in
% \end{figure}

\subsection{Continuous Case}
%\prasanna{Figure 5 snd 6 are taking too much whitespace, can the font size be made consistent with other figures.} \joshua{I will look into it. Right now, Figures 1-4 are generated by Yuheng in MATLAB, and Figures 5,6 are generated by me using Python. It would be best to have figures generated by the same program, so I'll look into copying my results into MATLAB.}
In the continuous case, we experiment on the Communities and Crimes (C\&C) dataset\footnote{http://archive.ics.uci.edu/ml/datasets/communities+and+crime}. The goal is to predict the crime rate $Y$ of a community given a set of 121 statistics $X$ (distributions of income, age, urban/rural, etc.). The 122-th statistic (percentage of black people in the community) is used as the sensitive variable $D$. All variables in this dataset are real-valued.
The dataset was split into 1794 training and 200 test samples. Following \cite{mary2019fairness}, we use a Neural Net with a 50-node hidden layer (which we denote as $f(x)$) and train a predictor $\hat{y} = T(f(x))$ with the mean squared error (MSE) loss and the soft-HGR penalty, varying $\lambda$. For soft-HGR, we use two 2-layer NNs with scalar outputs as the two maximal correlation functions $\bg$ and $\bh$, and trained them according to \eqref{eq:cts_ind_loss} (independence) or \eqref{eq:cts_sep_loss} (separation). We then computed the test MSE and test ``discrimination'' in each case.

For independence, our metric was $I(\hat{Y};D)$, approximated using the $\HGR_{\inf}$ method of \cite{mary2019fairness} as well as with a standard \kNN-based mutual information estimator \cite{gao2018demystifying}. For separation, we computed $I(\hat{Y};D|Y)$ using the two methods method.
We report the results of our experiment as well as that of the $\chi^2$ method of \cite{mary2019fairness} with the same architecture. The results of the separation experiment is presented in Figures \ref{cts-sep}. Results for independence can be found in the supplementary materials.

\begin{figure}[!t]
%\vskip 0.2in
\centering
\begin{minipage}{\linewidth}
\centerline{\includegraphics[width=\linewidth]{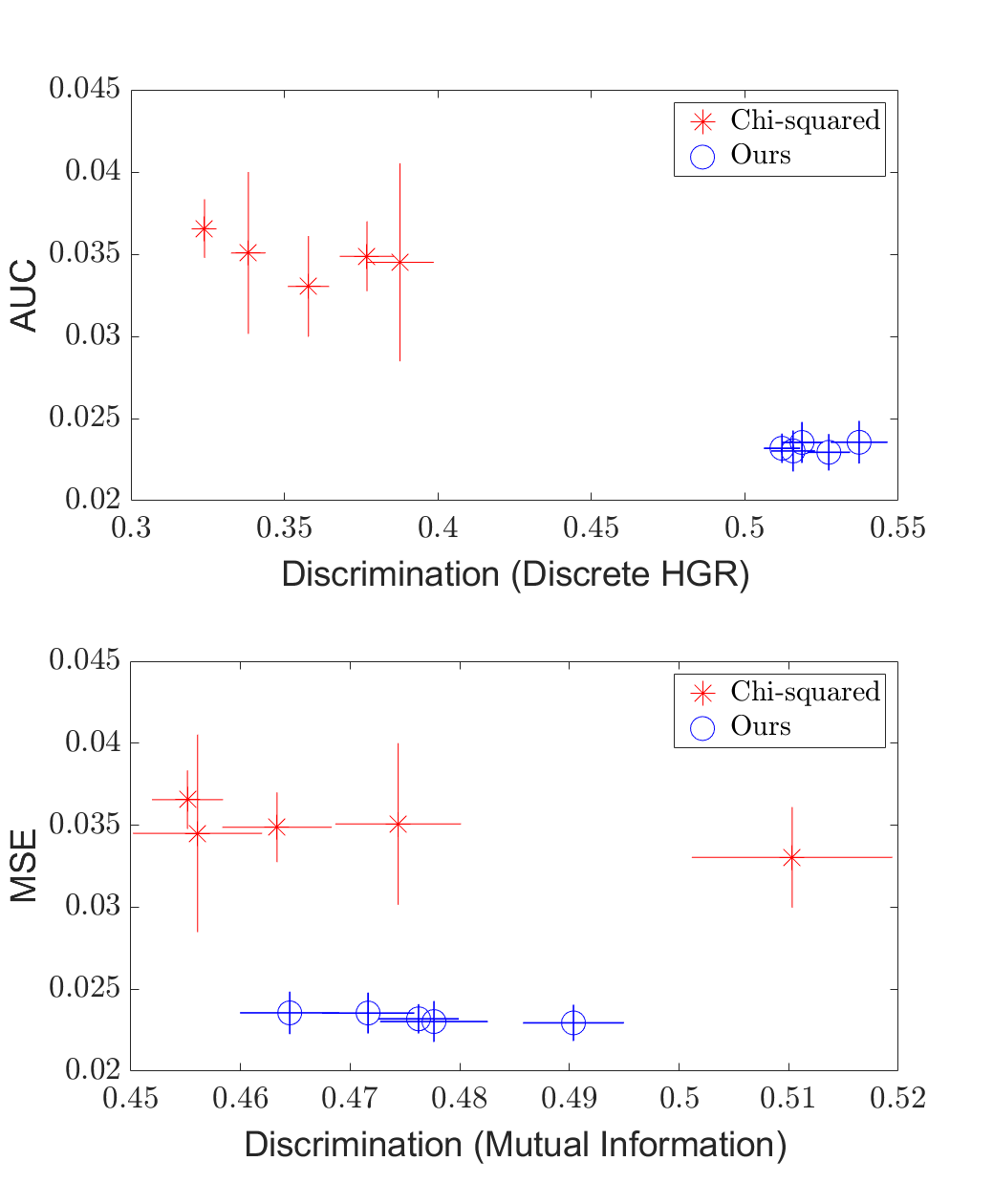}}
\end{minipage}
\caption{Separation regularization on the C\&C dataset, with MSE plotted against $\HGR_{\inf}(\hat{Y},D|Y)$ (Top) and $I(\hat{Y};D|Y)$ (Bottom), respectively.}\label{cts-sep}
\end{figure}   

\begin{figure}[!t]
%\vskip 0.2in
\centering
\begin{minipage}{\linewidth}
\centerline{\includegraphics[width=\linewidth]{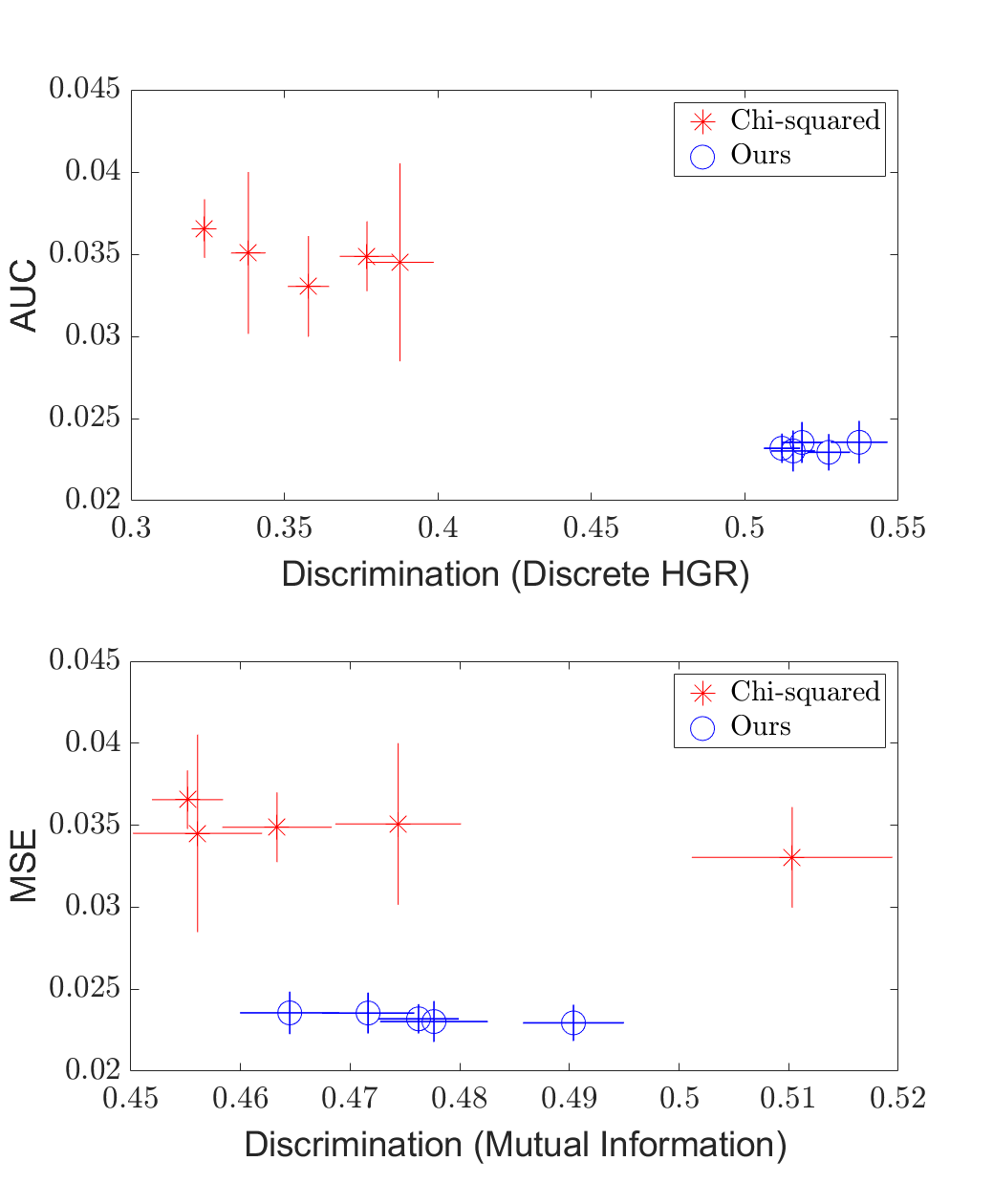}}
\end{minipage}
\caption{Separation regularization on the C\&C dataset in the \textit{few-shot} settings, with MSE plotted against $\HGR_{\inf}(\hat{Y},D|Y)$ (Top) and $I(\hat{Y};D|Y)$ (Bottom), respectively.}\label{cts-sep-few}
\end{figure}   

As expected, we see a tradeoff between the MSE and discrimination, creating a frontier of possible values. We also see that the Soft-HGR penalty does not outperform the $\chi^2$ method when measuring discrimination using $\HGR_{\inf}$, but does provide modest gains when using mutual information, thus revealing the sensitivity of these results to the choice of fairness measure while also showing that the Soft-HGR method can be superior in performance to the $\chi^2$ method under the right conditions.

Moreover, our method runs significantly faster than the $\chi^2$ method (on the order of seconds per iteration for our method versus just under a minute per iteration for the comparison method), as the $\chi^2$ method requires computation over a mesh grid of a Gaussian KDE, which scales with the product of the number of ``bins'' (mesh points) and the number of training samples, while our method only scales with the number of samples ($O(n)$). For large bandwidths, $d$ can become quite large. KDE methods also scale poorly with dimensionality \cite{wang2019nonparametric} in an exponential manner, and thus if $d$ is high-dimensional, the $\chi^2$ method would run much slower than our method, which can take in an arbitrarily-sized input and scale linearly with the dimensionality of the input multiplied by the number of samples. Empirically, we find that our method runs around five times faster.

We also run experiments to illustrate how our method's simplicity allows it to adapt to the few-shot, few-epoch regime faster than that of the $\chi^2$ method. We take 10 ``few-shot'' samples from the training set, then train a network to predict $Y$ from $X$ without any fairness regularizer using the full training set. Then, we run 5 more iterations of gradient descent on the trained model using the fairness-regularized objective and the 10 few-shot samples, and compare the separation results between the Soft-HGR and $\chi^2$ regularizer. We choose to compare to the $\chi^2$ regularizer as it is one of the few methods designed to handle continuous $D$. The results are shown in Figure \ref{cts-sep-few} (Independence results can be found in the supplementary materials). Once again, we see the tradeoff curve, and see our method outperform the $\chi^2$ method under both fairness measures, and that it appears to be competitive with the standard case in just a few iterations, while the $\chi^2$ method is still far from achieving the original MSE. Thus, in situations where, due to ethical/legal issues, only a few samples labeled with the sensitive attribute can be collected, fairness can still be enforced.

\section{Conclusions}

As machine learning algorithms gain more relevance, more focus will be placed upon ensuring their fairness. We have presented a framework using the HGR maximal correlation which provides effective and computationally efficient methods for enforcing independence and separation constraints, and derived algorithms for fair learning on discrete and continuous data which provide competitive tradeoff curves. In addition, we have also shown promising results in the few-shot setting, and suggested a method for rapidly adapting a classifier to improve fairness. In the future, it would be beneficial to extend this framework to other criteria (e.g. sufficiency), and to to determine how to use this framework to enforce fairness in a transfer learning setup coupled with the few-shot setting, to determine how to fairly adapt a classifier to a new task.

\section*{Broader Impact}

It is of utmost importance to develop meaningful measures of fairness as well as flexible frameworks to adapt to changing needs in society. With the rapid pace of development in AI, as well as evolving social awareness of systematic discrimination against protected groups, the ability to enforce fairness in any form it might take in the future is vital in ensuring that these systems continue to be used to real world applications to prevent harm to the most vulnerable in society. Furthermore, having a conceptually-grounded framework for thinking about fairness can also help inspire legal precedents, which are currently very ad-hoc in terms of how they define discrimination mathematically.

Unfortunately, these methods are also fraught with risk if misused. Applying the wrong criterion to a problem can result in unexpected discriminatory practice (e.g. the criticisms of Affirmative Action, which is derived from Independence). Thus, care must be taken when using these techniques, but ultimately they can be used to create a more just society.

\bibliography{aaai.bib}

\end{document}